# PHOCNet: A Deep Convolutional Neural Network for Word Spotting in Handwritten Documents

Sebastian Sudholt, Gernot A. Fink
Department of Computer Science
TU Dortmund University
44221 Dortmund, Germany
Email: {sebastian.sudholt, gernot.fink}@tu-dortmund.de

*Abstract*—In recent years, deep convolutional neural networks have achieved state of the art performance in various computer vision task such as classification, detection or segmentation. Due to their outstanding performance, CNNs are more and more used in the field of document image analysis as well. In this work, we present a CNN architecture that is trained with the recently proposed PHOC representation. We show empirically that our CNN architecture is able to outperform state of the art results for various word spotting benchmarks while exhibiting short training and test times.

## I. INTRODUCTION

In recent years, Convolutional Neural Networks (CNN) have received increased attention as they are able to consistently outperform other approaches in virtually all fields of computer vision. Due to their impressive performance, CNNs have found their way into document image analysis as well. However, the use of CNNs in word spotting applications has been scarce. Word spotting is an effective paradigm to index document images for which a direct classification approach would be infeasable. In [1] the authors use a pretrained CNN to perform word spotting on the IAM database. However, this approach has several short comings: Each word image has to be cropped to a unit width and height which almost always distorts the image. Moreover, their CNN is pretrained on the ImageNet database which comes from a completely different domain than word images. Although the results are good, this approach bares the question whether a CNN trained on word images only can produce even better results.

In this work, we present a CNN architecture specifically designed for word spotting. By using the recently proposed *Pyramidal Histogram of Characters (PHOC)* [2] as labels, this CNN is able to achieve state-of-the-art performance in Query-by-Example as well as Query-by-String scenarios on different datasets. This is also due to the network being able to accept input images of arbitrary size. Figure 1 gives a brief overview over our proposed approach. Due to the PHOCs being used for training, we refer to our deep CNN as *PHOCNet* throughout the paper.

## II. RELATED WORK

### A. Word Spotting

Word spotting has gained major attention ever since it was first proposed in [3]. The goal in word spotting is to retrieve word images from a document image collection which are relevant with respect to a certain query. This paradigm has shown itself to be very effective in situations where a recognition approach does not produce reliable results.

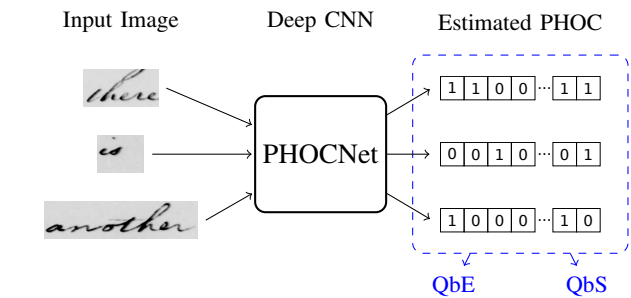

Fig. 1. Overview over the proposed PHOCNet approach for Query-by-Example (QbE) and Query-by-String (QbS) word spotting.

Numerous query representations have been proposed throughout the literature. In *Query-by-Example (QbE)* word spotting, e.g. [2]–[4], the query is a word image and retrieval is based on the visual similarity of the test word images. This approach, however, poses certain limitations in practical applications as the user has to identify a query word image from the document image collection. This might either already solve the task (does the collection contain the query?) or be tedious when looking for infrequent words as queries [5], [6].

Thus the focus for word spotting has shifted towards *Query-by-String (QbS)* based approaches [2], [5], [7]. Here, the user supplies the word spotting system with a textual representation of the sought word and is returned a list of word images. The drawback of QbS systems with respect to QbE systems is that they need to learn a model to map from textual representation to image representation first, thus requiring annotated word images.

The predominant features used in word spotting have been SIFT descriptors [2], [4], [5], [7], [8], geometric features [3], [9] and HOG-based descriptors [10]. All features share their expert design and the ability to be generated in an unsupervised fashion. For other computer vision tasks, these so called shallow features have been outperformed by features learned in a supervised manner through deep neural network architectures.

*B. Convolutional Neural Networks*

Although CNNs were initially proposed in the early 1990's [11], it has only been recently that they received major attention. The advent of large scale datasets such as ImageNet [12] and highly optimized implementations running on graphic cards enabled the typically thousands of parameters of such a network to be trained in an acceptable amount of time. Krizhevsky et al. [13] were the first to use CNNs in the *ImageNet Large Scale Visual Recognition Challenge*, largely outperforming other approaches. This competition has been ruled by CNNs ever since with the winning teams always featuring "very deep" architectures [14], [15]

Despite their large success, there has been very limited work on using CNNs for word spotting. In [1] a pretrained deep CNN is finetuned to learn classes of word images. The output is then used to perform word spotting. However, using a pretrained CNN and finetuning on word images might leave the network stuck in a local optimum specific to the initial training domain (in this case the ImageNet database) which might not yield top performance. Additionally, the CNN used needs a fixed image size. The majority of word images has thus either to be scaled or cropped to this size. This leads to either distorting or erasing important parts of the word image. In our approach, the word image size is not altered which helps the CNN to generalize better over common semantical units (i.e. characters, bigrams,...).

The approach closest to ours is described in [16]. Here, an ensemble of a character and an n-gram CNN is used to perform unconstrained text recognition. While the first CNN predicts the character at each position of a word image the latter classifies whether a certain n-gram is present in the word. This approach resizes the word images similar to [1]. However, the encoding of the characters at the individual positions is somewhat similar to the PHOC representation. The only difference is that, while the representation in [16] can only deal with words of up to 23 characters, the PHOC representation can handle arbitrary word lengths.

## III. METHOD

*A. CNN Elements*

CNN architectures can generally be split into two parts. The first is the convolutional part that usually constitutes of convolutional and pooling layers. Convolutional layers consist of a number of so called *filters* with which the input image is convolved. The output is a number of *feature maps* which can be the input to another layer of the CNN. Each feature map is produced by applying one of the filters in the respective convolution layer to the input. In order to introduce non-linearity into CNNs, the output of convolutional layers is passed through an activation function. Traditionally, the activation function of choice has been the sigmoid function

$$sg(x) = \frac{1}{1 + e^{-x}} \quad (1)$$

as it is easily differentiable. However, this function leads to training stalling for deep neural networks due to the *Vanishing Gradient Problem* [17]. In order to circumvent this problem, state-of-the-art CNN architectures have made use of the *Rectified Linear Unit* $r(x) = \max(0, x)$ as non-linear activation function [13]. After applying the activation function the receptive field size can be expanded by using *Pooling layers*. These CNN layers aggregate filter repsonses by downsampling the feature map. The predominant pooling strategy in deep CNNs has been *Max Pooling*. In Max Pooling, the filter responses over a certain local region (i.e. receptive field) are taken and only the maximum filter response is passed to the next layer.

The convolutional part of a CNN can be thought of as producing a feature representation that can be fitted to the data at hand in a supervised manner. After this part, deep CNNs usually make use of a standard *Multilayer Perceptron (MLP)* as a classifier. Here, multiple so called fully connected layers are stacked together to form the MLP.

In usual single label image classification tasks, training a CNN is carried out by first applying the softmax function

$$sm(\mathbf{o})_i = \frac{e^{o_i}}{\sum_{j=1}^{n} e^{o_j}} = \hat{y}_i \quad (2)$$

to the output $\mathbf{o}$ of the last layer of the CNN in order to generate the output vector $\hat{\mathbf{y}}$ of predicted pseudo class probabilities (see figure 3). This can be seen as adding a special non-linear scaling layer to the CNN. In order to adapt the parameters to the data, the cross entropy loss $l$ between the one-hot encoded label vector $\mathbf{y}$ and $\hat{\mathbf{y}}$ is computed as

$$l(\mathbf{y}, \hat{\mathbf{y}}) = -\frac{1}{n} \sum_{i=1}^{n} \left[ y_i \log \hat{y}_i + (1 - y_i) \log(1 - \hat{y}_i) \right]. \quad (3)$$

The error is then backpropagated through the network.

*B. PHOCNet Architecture*

The architecture of our PHOCNet is visualized in figure 2. The design choice is based on a number of considerations. First, we only use $3 \times 3$-convolutions followed by *Rectified Linear Units (ReLU)* in the convolutional parts of the neural network. These convolutions have been shown to achieve better results compared to those with a bigger receptive field as they impose a regularization on the filter kernels [14]. Similar to the design presented in [14], we select a low number of filters in the lower layers and an increasing number in the higher layers. This leads to the neural network learning fewer features for smaller receptive fields and more features for higher level and thus more abstract features.

Usually, CNNs are fed with images of the same width and height. Most word images would thus have to be either cropped or rescaled. As was already mentioned in section II-B, resizing might severely distort similar semantic aspects in the visual domain (consider the character *a* in two hypothetical word images showing *as* and *about*). In [18], the authors present a form of pooling layer called *Spatial Pyramid Pooling*. This type of layer allows CNNs to accept differently sized input images and still produce a constant output size which is essential for training the network. The key insight is, that convolutional and pooling layers can already handle different image sizes as the only thing changing is the feature map size. In traditional CNN architectures, only fully connected layers can not deal with changing image sizes. Thus the authors propose to use a pooling strategy similar to the well known spatial pyramid principal as the last pooling layer before the fully connected part of the CNN. This way, a CNN can be fed with arbitrarly sized input images and is still able to produce a constant output size. In our method, we use a 3-level Spatial Pyramid max pooling to circumvent the need for cropping or resizing the input image.

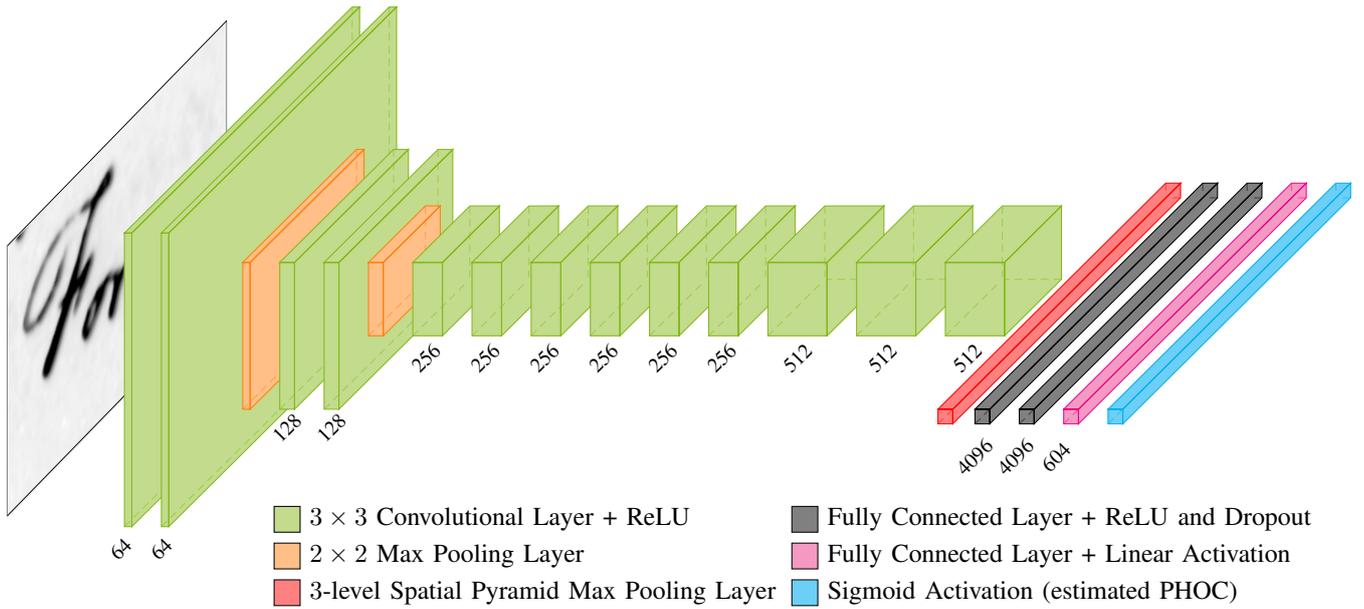

Fig. 2. The figure displays the PHOCNet architecture. All convolutional layers are depicted by a green, all max pooling layers by an orange and the fully connected layers by a black box. The red box depicts the spatial pyramid pooling layer while the blue box represents the sigmoid activation layer. The number of filters for each convolutional layer is shown underneath as are the number of neurons for the fully connected layers. The number of neurons in the last layer is equal to the size of the PHOC. Convolutional layers use stride 1 and apply 1 pixel padding. Pooling layers use stride 2. If the PHOC is created using only the lower case characters from the Latin alphabet plus the ten digits, it has a dimensionality of 604.

For a word spotting task, using the single label classification paradigm as presented in section III-A is infeasable due to a number of reasons: If the query word class is not among the training classes (out of vocabulary), it is not obvious how to perform QbE word spotting. Even worse, QbS word spotting is altogether impossible for these queries. Also, the softmax layer is usually overconfident for misclassifications which makes it hard to automatically detect misclassifications.

In order to alleviate the problems at hand, we make use of the recently proposed PHOC representation [2]. A PHOC is a binary pyramidal representation of a character string. It encodes visual attributes of the corresponding word image. Here, an attribute refers to a semantic unit that may be shared between word images. Intuitive attributes of a word image are its characters. The PHOC encodes if a certain attribute (i.e. character) is present in a certain split of the string representation of a word. For exmaple, the 2nd level of the PHOC encodes whether the word contains a certain character in the first or second half of the word. In [2] the authors skip a global representation and represent a word image by a PHOC with $2, 3, 4$ and $5$ splits. This yields a binary histogram of size $504$. Additionally, they use the $50$ most frequent bigrams at level $2$. Using the lower case Latin alphabet plus the ten digits, the PHOC has a size of $604$. The PHOC allows to transfer knowledge about attributes from the training images to the test images as long as all attributes in the test images are present in the training images.

The output of the resulting deep CNN can be used as a holistic word image representation in a simple retrieval approach. For QbE, the representations can be compared directly while for QbS a PHOC can be generated from the query and be compared to the output representation of the neural network.

In order to train a deep CNN with PHOCs, the softmax layer can no longer be used as only one element in the training vector is 1 whereas multiple elements of the PHOC label can be 1. However, training the CNN with PHOCs as labels can be seen as a multi-label classification task. Thus, we swap the softmax function by a sigmoid activation function (equation 1) which is applied to every element of the output vector. Figure 3 visualizes the changes compared to a standard softmax CNN. In this figure and also in figure 2 we show the sigmoid activation as a seperate layer in order to visualize the replacement of the softmax layer. Here, $\hat{a}_i$ refers to the pseudo probability for attribute $i$ being present in the word image. This way, each attribute is interpreted as a label in a multi-label classification task. For training, we apply the cross entropy loss (equation 3) as is done in single label classification tasks (softmax classification) and backpropagate the error through the CNN.

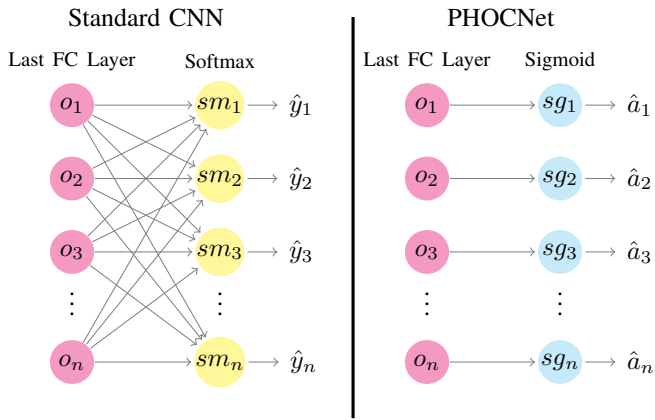

Fig. 3. Visualization of a standard softmax output and the output of the PHOCNet.

After training, the PHOCNet outputs an estimated PHOC for a given input image. This output can now be used in a simple nearest neighbor approach in order to perform either QbE or QbS word spotting. Note that the last layer of sigmoid activations can be viewed as being similar to the Platt's scaling applied to the output of the Attribute SVMs in [2].

### C. Regularization

The vast amount of parameters in our PHOCNet makes it prone to overfitting. Hence, we apply a number of regularization techniques that have become common when using deep CNNs.

In many image classification tasks, CNNs have greatly benefitted from the use of *Dropout* in the fully connected layers [13], [14]. In Dropout, activations of a certain layer are randomly set to 0 [19]. This acts as a regularizer on the CNN as neurons following a layer with dropout can no longer rely on a neuron in the previous layer to be active for a specific input image. In our approach, we apply Dropout of 0.5 to all but the last fully connected layer (all black layers in figure 2).

Additionally, we augment the set of training word images. This balances the classes of word images and imposes another measure of regularization on the CNN. For the augmentation we randomly sample a number of word images from each class and apply a random affine transform per sampled word image. This affine transform is obtained by selecting the relative coordinates $(0.5, 0.3), (0.3, 0.6)$ and $(0.6, 0.6)$ and multiplying each coordinate value with a random factor drawn from a uniform distribution with limits $[0.8, 1.1]$. The transform is then the homography needed to obtain the new coordinates from the initial coordinates. For each class we generate images such that the classes are balanced and the number of training images amounts to $500\,000$.

### D. Training

We train our PHOCNet using stochastic gradient descent with a batch size of 10, momentum of 0.9, weight decay of $5 \cdot 10^{-5}$ and an initial learning rate of $10^{-4}$. The selection of these parameters is based on those used in [14] for a similar network architecture. Training is run for $80\,000$ iterations with the learning rate being divided once by 10 after $70\,000$ iterations. The chosen parameters stay the same for all experiments.

Initializing the CNN parameters before training is a critical aspect in learning the model. We follow [20] and initialize the weights by randomly sampling from a zero-mean uniform distribution with variance $\frac{2}{n}$ where $n$ is the number of parameters in a given layer. Likewise, layer biases are initialized with 0. We found this initialization to produce slightly better results compared to initializing from a Gaussian distribution as proposed in [21]. Training is then carried out on a single Nvidia GeForce Titan X GPU using the Caffe framework [22].

## IV. EXPERIMENTS

### A. Datasets

We use a total of four datasets to assess the performance of the PHOCNet. The first is the well known **George Washington dataset (GW)** which has become a standard benchmark in word spotting. It consists of 20 pages of correspondences from George Washington and his associates which contain a total of 4860 words. Due to the homogeneous writing style it is considered a single-writer dataset. As there is no official partition in training and test images, we use the approach as was presented in [2] and perform a fourfold cross validation. We use the exact same partitions as were used in [2][1].

The second dataset is the **IAM Handwritten Database (IAM)**[2]. It is made up of 115 320 words written by 657 writers. We use the official partition available for writer independent text line recognition. In order to be able to directly compare our results to [2] we exclude the official stop words as queries but keep them as distractors in the dataset.

The third dataset is the **Esposalles database** [24][3]. It is an ancient marriage license register written between 1451 and 1905 by multiple writers. Here, we use the official word partition which contains 32 052 training images and 12 048 test images.

The last dataset used is the **IFN/ENIT database**[4]. Different from the previous datasets it features Arabic script in the form of handwritten city names. The IFNENIT is made up of seven different subsets. We use the common partition of subsets $a, b$ and $c$ for training and subset $d$ for testing. This way, the training set contains a total of 19 724 word images while the test set contains 6735 images. In order to extract PHOCs from the Arabic script we used a reduced character set which was created in the following way: First all character shapes were mapped to their representative Arabic characters. Characters with optional *Shadda* diacritic are replaced with characters without the Shadda diacritic. Special two-character-shape ligature models were mapped to two-character ligature models without the shape contexts. This mapping produces a character set of size 50, the corresponding PHOC representation has a dimensionality of 800.

---
[1]partitions available at https://github.com/almazan/watts/tree/master/data
[2]http://www.iam.unibe.ch/fki/databases/iam-handwriting-database
[3]http://dag.cvc.uab.es/dag/?page_id=3704
[4]http://www.ifnenit.com/download.htm

TABLE I
RESULTS FOR THE QBE AND QBS EXPERIMENTS IN MAP [%]

| Method | GW | | IAM | | Esposalles | | IFN/ENIT | |
|---|---|---|---|---|---|---|---|---|
| | QbE | QbS | QbE | QbS | QbE | QbS | QbE | QbS |
| BLSTM (*) [9] | - | 84.00 | - | 78.00 | - | - | - | - |
| SC-HMM (*) [23] | 53.10 | - | - | - | - | - | 41.60 | - |
| LSA Embedding [5] | - | 56.54 | - | - | - | - | - | - |
| Finetuned CNN [1] | - | - | 46.53 | - | - | - | - | - |
| Attribute SVM [2] | 93.04 | 91.29 | 55.73 | 73.72 | - | - | - | - |
| Softmax CNN | 78.24 | - | 48.67 | - | 89.38 | - | 91.78 | - |
| PHOCNet | **96.71** | **92.64** | **72.51** | **82.97** | **97.24** | **93.29** | **96.11** | **92.14** |

### B. Protocol

We evaluate our PHOCNet in segmentation-based QbE and QbS scenarios. For both scenarios we use the same protocol as was presented in [2]: First, the ground truth bounding box is used to create a perfect segmentation. Then the PHOCNet is trained on the training partition of each dataset (for training parameters see section III-D). During query time, each word image in the test set is used once as a query to rank the remaining word images in the test set for QbE. As a distance measure, we chose the Bray-Curtis dissimilarity [8]. Queries which appear only once in the test set are discarded (they still appear as distractors in the other retrieval lists though). For QbS we extract all unique transcriptions in the test set and use their PHOC representation as queries to rank all images in the test set. As a performance measure, the *Mean Average Precision (mAP)* is calculated for all queries (only valid queries for QbE).

We compare the performance of our PHOCNet to state-of-the-art results reported in the literature. As an additional baseline, we evaluate a deep CNN trained to predict word labels on the four datasets as well (Softmax CNN). This CNN has the same architecture as the PHOCNet except for using a softmax activation instead of a sigmoid activation as the last layer (see figure 3). During some pre-experiments, it became evident that the Softmax CNN needs considerably more training iterations than the PHOCNet. Thus, we set the total number of iterations for the Softmax CNN to 500 000 with the learning rate being divided by 10 after 250 000 iterations.

### C. Results

Table I lists the results for the different experiments run on the four datasets. Methods marked with an asterisk do not share the same evaluation protocol and can thus not be compared to our method directly. However, we include them to give a general idea on where the PHOCNet ranks performance-wise. For example, in [9] the authors retrieve entire lines of word images which in [2] could be shown to be easier than retrieving single word images as is done here. In [23], a fifefold cross validation is performed which leaves the system with a smaller test set and thus also an easier retrieval task.

Figure 4 displays the mAP over the course of the training for the four QbE experiments. Note that an iteration means computing the gradient for the current batch and adjusting the weights of the CNN accordingly.

### D. Discussion

There are a number of interesting observations to make from the experiments. First, we can disprove the notion that deep CNNs always need massive amounts of training data when trained from scratch as is stated in [1]. Using simple data augmentation and common regularization techniques, we are able to outperform other methods on even small datasets like the GW (in our setup 3645 training images, 964 classes in training on average). Driven by this result, we investigated using even smaller training partitions for this dataset. Using the same cross validation splits as presented in section IV-A and taking only one fold for training and one fold for testing, the PHOCNet was able to achieve a mAP of 86.59 (1215 training images, 488 classes in training on average).

Second, the multi-label classification approach in our PHOCNet leads to faster training times and higher performance compared to a standard softmax CNN. For the IAM, training terminates in less than 17 hours. Estimating the PHOC representation for a given word image takes less than 28 ms. In comparison, training Attribute SVMs on the IAM database takes roughly two days [2]. Moreover, if training speed is the primary concern, highly competitive results can already be achieved after 40 000 iterations (see figure 4).

Another very appealing aspect of our PHOCNet is its robustness with respect to the parametrization. In all experiments, we chose the exact same set of parameters. Additionally, the PHOCNet's performance on the IAM database shows its robustness in a multi writer scenario.

## V. CONCLUSION

In this paper we introduced PHOCNet, a deep CNN architecture designed for word spotting. It is able to process input images of arbitrary size and predicts the corresponding PHOC representation. We show empirically that the PHOCNet is able to outperform current state-of-the-art approaches on various datasets. Compared to a CNN trained on the ImageNet database and finetuned on word images, it is able to produce vastly better results [1]. Likewise, it is able to outperform Attribute SVMs in both Query-by-Example and Query-by-String scenarios on the presented datasets. This holds true for Latin as well as Arabic script.

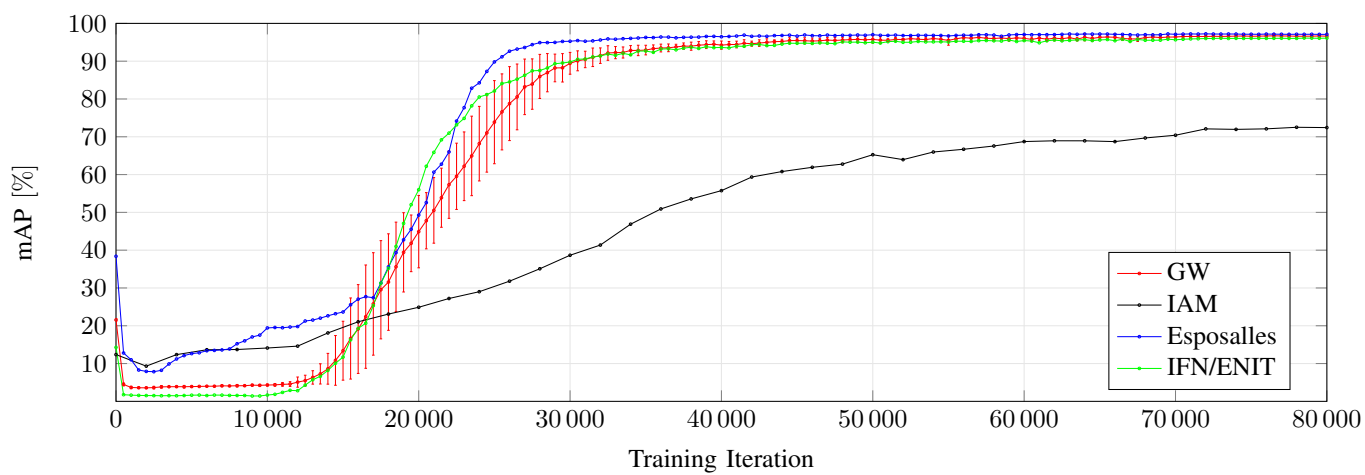

Fig. 4. The figure displays the mAP over the different training iterations for the four QbE experiments (GW showing cross validation standard error).

ACKNOWLEDGMENT

The authors thank Irfan Ahmad for helping to set up the IFN/ENIT experiment and supplying the character mapping.
REFERENCES

[1] A. Sharma and K. Pramod Sankar, "Adapting off-the-shelf CNNs for Word Spotting & Recognition," in *International Conference on Document Analysis and Recognition*, 2015, pp. 986–990.
[2] J. Almazán, A. Gordo, A. Fornés, and E. Valveny, "Word Spotting and Recognition with Embedded Attributes," *Transactions on Pattern Analysis and Machine Intelligence*, vol. 36, no. 12, pp. 2552–2566, 2014.
[3] T. M. Rath and R. Manmatha, "Word Spotting for Historical Documents," *IJDAR*, vol. 9, pp. 139–152, 2007.
[4] M. Rusiñol, D. Aldavert, R. Toledo, and J. Lladós, "Efficient segmentation-free keyword spotting in historical document collections," *Pattern Recognition*, vol. 48, no. 2, pp. 545–555, 2015.
[5] D. Aldavert, M. Rusinol, R. Toledo, and J. Llados, "Integrating Visual and Textual Cues for Query-by-String Word Spotting," in *International Conference on Document Analysis and Recognition*, 2013, pp. 511–515.
[6] M. Rusiñol, D. Aldavert, R. Toledo, and J. Lladós, "Towards Query-by-Speech Handwritten Keyword Spotting," in *International Conference on Document Analysis and Recognition*, 2015, pp. 501–505.
[7] L. Rothacker and G. A. Fink, "Segmentation-free query-by-string word spotting with bag-of-features HMMs," in *International Conference on Document Analysis and Recognition*, Nancy, France, 2015.
[8] S. Sudholt and G. A. Fink, "A Modified Isomap Approach to Manifold Learning in Word Spotting," in *37th German Conference on Pattern Recognition*, ser. Lecture Notes in Computer Science, Aachen, Germany, 2015.
[9] V. Frinken, A. Fischer, R. Manmatha, and H. Bunke, "A Novel Word Spotting Method Based on Recurrent Neural Networks," *Transactions on Pattern Analysis and Machine Intelligence*, vol. 34, pp. 211–224, 2012.
[10] J. Almazán, A. Fornés, and E. Valveny, "Deformable HOG-Based Shape Descriptor," in *International Conference on Document Analysis and Recognition, ICDAR*, 2013, pp. 1022–1026.
[11] Y. LeCun, B. Boser, J. S. Denker, D. Henderson, R. E. Howard, W. Hubbard, and L. D. Jackel, "Handwritten Digit Recognition with a Back-Propagation Network," *Advances In Neural Information Processing Systems*, pp. 396–404, 1990.
[12] O. Russakovsky, J. Deng, H. Su, J. Krause, S. Satheesh, S. Ma, Z. Huang, A. Karpathy, A. Khosla, M. Bernstein, A. C. Berg, and L. Fei-Fei, "ImageNet Large Scale Visual Recognition Challenge," *International Journal on Computer Vision*, vol. 115, no. 3, pp. 211–252, 2015.
[13] A. Krizhevsky, I. Sutskever, and G. E. Hinton, "ImageNet Classification with Deep Convolutional Neural Networks," *Advances In Neural Information Processing Systems*, pp. 1097–1105, 2012.
[14] K. Simonyan and A. Zisserman, "Very Deep Convolutional Networks for Large-Scale Image Recognition," *arXiv*, pp. 1–13, 2014.
[15] C. Szegedy, W. Liu, Y. Jia, P. Sermanet, S. Reed, D. Anguelov, D. Erhan, V. Vanhoucke, and A. Rabinovich, "Going Deeper with Convolutions," *arXiv preprint arXiv:1409.4842*, pp. 1–12, 2014.
[16] M. Jaderberg, K. Simonyan, A. Vedaldi, and A. Zisserman, "Deep Structured Output Learning for Unconstrained Text Recognition," in *International Conference on Learning Representations*, 2015, pp. 1–10.
[17] R. Pascanu, T. Mikolov, and Y. Bengio, "On the difficulty of training recurrent neural networks," in *ICML*, no. 2, 2013, pp. 1310–1318.
[18] K. He, X. Zhang, S. Ren, and J. Sun, "Spatial Pyramid Pooling in Deep Convolutional Networks for Visual Recognition," *European Conference on Computer Vision*, pp. 346–361, 2014.
[19] N. Srivastava, G. Hinton, A. Krizhevsky, I. Sutskever, and R. Salakhutdinov, "Dropout : A Simple Way to Prevent Neural Networks from Overfitting," *Journal of Machine Learning Research*, vol. 15, pp. 1929–1958, 2014.
[20] X. Glorot and Y. Bengio, "Understanding the Difficulty of Training Deep Feedforward Neural Networks," *AISTATS*, vol. 9, pp. 249–256, 2010.
[21] K. He, X. Zhang, S. Ren, and J. Sun, "Delving Deep into Rectifiers: Surpassing Human-Level Performance on ImageNet Classification," in *International Conference on Computer Vision*, 2015, pp. 1026–1034.
[22] Y. Jia, E. Shelhamer, J. Donahue, S. Karayev, J. Long, R. Girshick, S. Guadarrama, and T. Darrell, "Caffe: Convolutional Architecture for Fast Feature Embedding," in *International Conference on Multimedia*, 2014, pp. 675–678.
[23] J. A. Rodríguez-Serrano and F. Perronnin, "A Model-Based Sequence Similarity with Application to Handwritten Word Spotting," *Transactions on Pattern Analysis and Machine Intelligence*, vol. 34, no. 11, pp. 2108–2120, 2012.
[24] V. Romero, A. Fornés, N. Serrano, J. A. Sánchez, A. H. Toselli, V. Frinken, E. Vidal, and J. Lladós, "The ESPOSALLES database: An ancient marriage license corpus for off-line handwriting recognition," *Pattern Recognition*, vol. 46, no. 6, pp. 1658–1669, 2013.